\documentclass[runningheads]{llncs}
\usepackage{graphicx}
\usepackage{amssymb}
\usepackage{latexsym}
\usepackage{stfloats}
\usepackage{graphicx}
\usepackage{hyperref}
\usepackage{url}
\usepackage{subfig}
\usepackage{amsmath}
\usepackage{slashbox}
\usepackage[table]{xcolor}
\usepackage{makecell}

\begin{document}
\title{EyeLoveGAN: Exploiting domain-shifts to boost network learning with cycleGANs}
\titlerunning{Domain-shifts to boost network learning with cycleGANs}
%
\author{Josefine Vilsb\o ll Sundgaard\thanks{These authors contributed equally to this work.} \and
Kristine Aavild Juhl$^\star$ \and
Jakob M\o lkj\ae r Slipsager}

\authorrunning{J. V. Sundgaard et al.}
%
\institute{Department of Applied Mathematics and Computer Science, Technical University of Denmark, Lyngby, Denmark}
\maketitle              
\begin{abstract}
This paper presents our contribution to the REFUGE challenge 2020. The challenge consisted of three tasks based on a dataset of retinal images: Segmentation of optic disc and cup, classification of glaucoma, and localization of fovea. We propose employing convolutional neural networks for all three tasks. Segmentation is performed using a U-Net, classification is performed by a pre-trained InceptionV3 network, and fovea detection is performed by employing stacked hour-glass for heatmap prediction. The challenge dataset contains images from three different data sources. To enhance performance, cycleGANs were utilized to create a domain-shift between the data sources. These cycleGANs move images across domains, thus creating artificial images which can be used for training.
\end{abstract}

\section{Introduction}
Glaucoma is a group of eye conditions that damage the optic nerve. It is one of the leading causes of irreversible, but preventable, blindness \cite{tham2014global}, and the incidence is expected to increase. The condition is typically caused by a high pressure in the eye, which damages the optic nerve with no warning signs. The condition of the retina, and thus the optical nerve, is examined using color fundus photography. This imaging technique is both economical and non-invasive. The Retinal Fundus Glaucoma Challenge (REFUGE2) \cite{orlando2020refuge} is a competition held as part of the Ophthalmic Medical Image Analysis (OMIA) workshop at the International Conference on Medical Image Computing and Computer Assisted Intervention (MICCAI) 2020. The goal of this challenge is to provide key tools for diagnosing glaucoma by releasing a large scale database. The challenge consists of three tasks: optic disc and cup segmentation, glaucoma classification, and fovea localization. The problem is challenged by the fact that the data is acquired from three different data sources. The training dataset is acquired with two different cameras, and the test dataset is acquired using a third camera. This paper presents our contribution to this challenge. Segmentation is performed using a U-Net, classification is performed by a pre-trained InceptionV3 network, and fovea detection is performed by employing stacked hour-glass for heatmap prediction of fovea location. To enhance performance and cope with the challenges of different data sources, cycleGANs are utilized to create a domain-shift between the data sources. These cycleGANs move images across domains, thus creating artificial images which can be used for training.

\section{Data}
\begin{figure*}[h!]
  \centering
  \subfloat[Domain 1]{\includegraphics[width=0.27\textwidth]{images/g0010.jpg}    \label{fig:dom1}}
  \hspace{0.05\textwidth}
  \subfloat[Domain 2]{\includegraphics[width=0.27\textwidth]{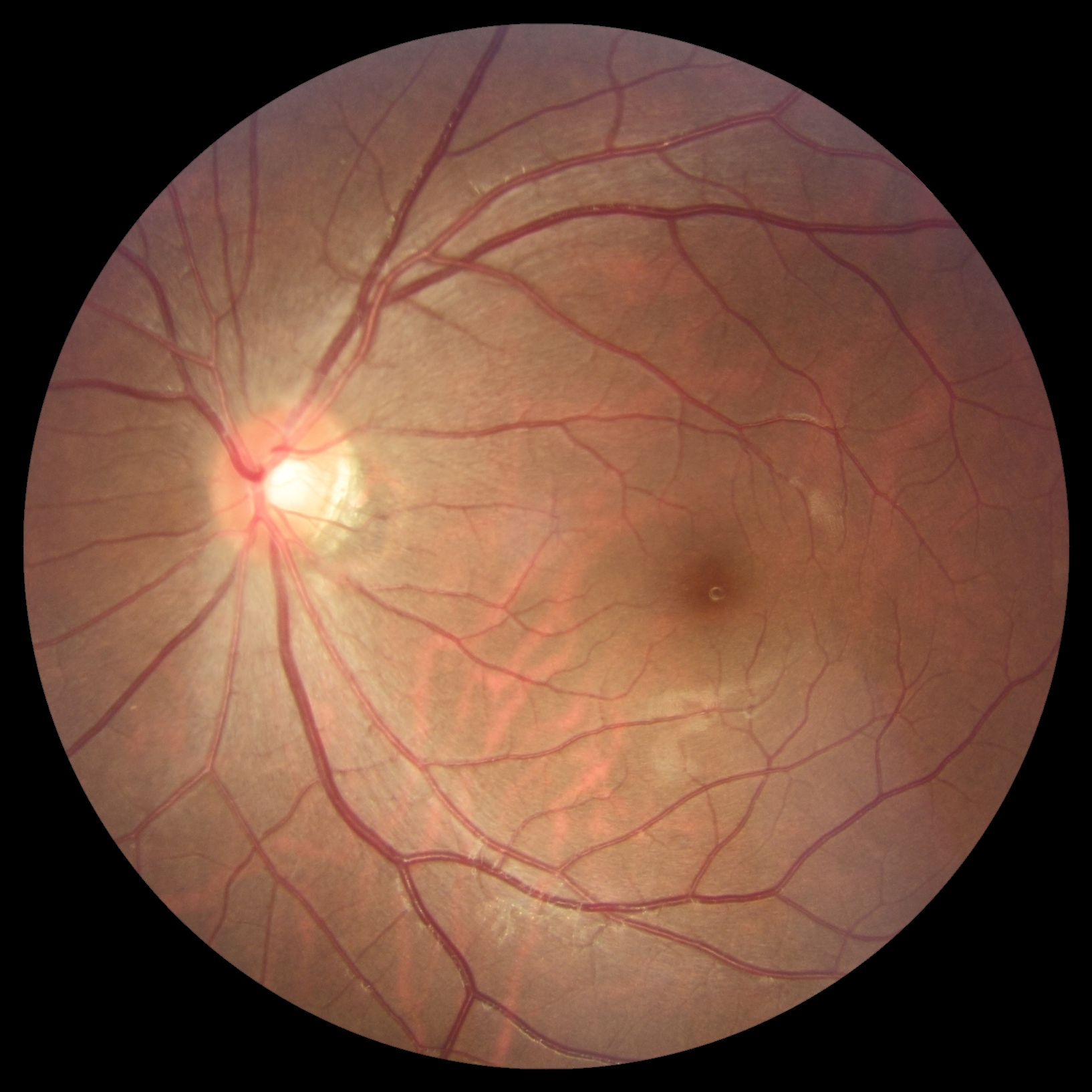}\label{fig:dom2}}
  \hspace{0.05\textwidth}
  \subfloat[Domain 3]{\includegraphics[width=0.27\textwidth]{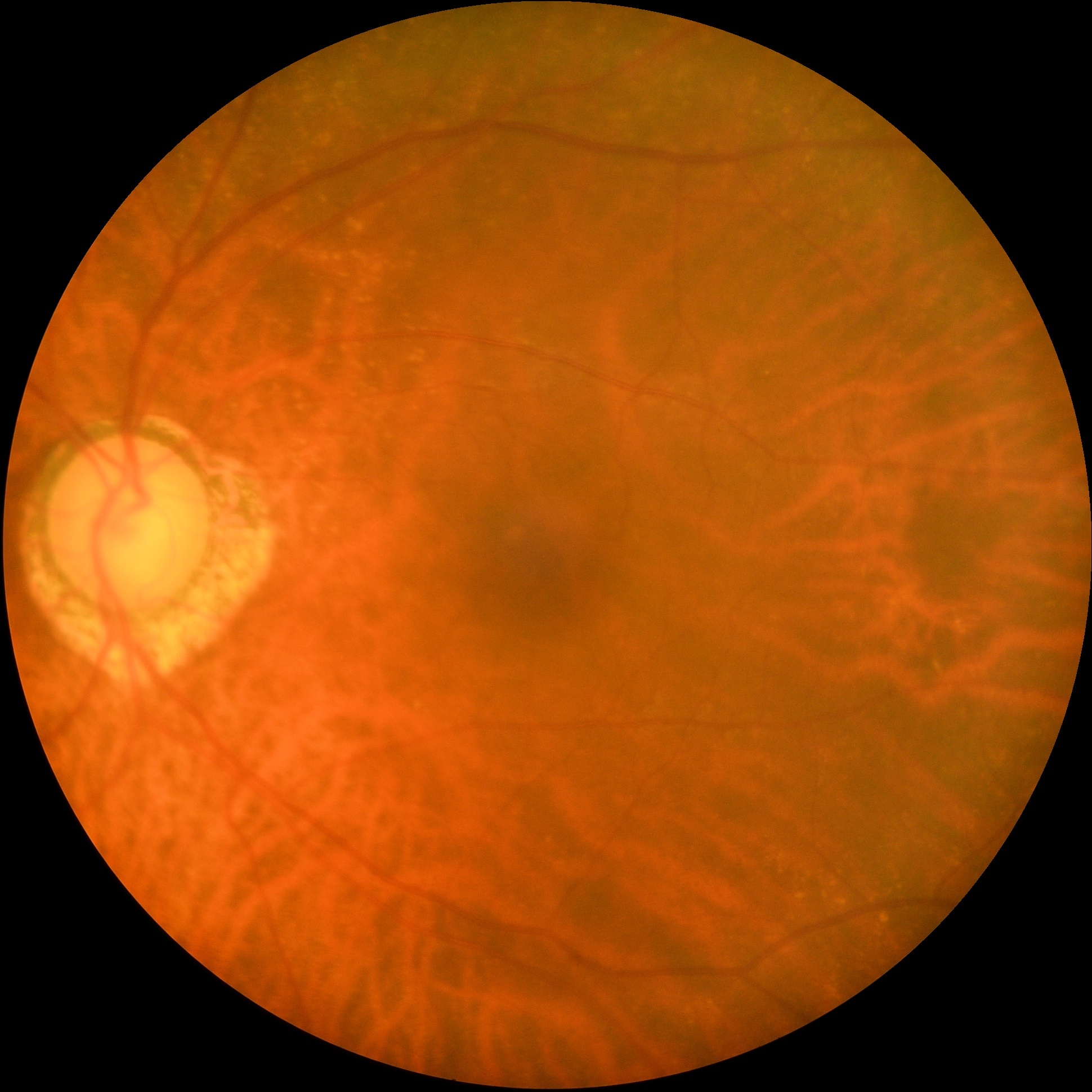}\label{fig:dom3}}
  \caption{Image examples from the three different domains in the challenge dataset}
  \label{fig:examples}
\end{figure*}

The approach is trained on the REFUGE challenge data consisting of 1200 annotated images from two different cameras (400 from one, 800 from another). The test dataset consists of 400 images from a third camera. The camera used for the first set of training images (later called domain 1) have the image dimensions 2124x2056, while the dimensions of the other part of the training dataset (later called domain 2) have the dimensions 1634x1634, and the test dataset (later called domain 3) has dimensions 1940x1940. 
Image examples from all three domains are seen in Figure \ref{fig:examples}.

For training of the various neural networks for this proposed method, the training data is split into training and validation using a 90/10 split. As the dataset is highly unbalanced with only 10\% of glaucoma cases, a class-wise stratified train-validation split was employed. The split was also generated to ensure that 10\% of images in each of the two training domains were used for validation. 

For classification of glaucoma and optic disc and cup segmentation, a region of interest is cropped out of the original images. The region is detected using a stacked hour-glass neural network for heatmap prediction of the center of the optic disc. This methods is explained in detail in Section \ref{sec:fovea}. A region of 500x500 pixels is cropped of the area of the optic nerve head, and this was used as training inputs.

\section{Methods}
As the dataset consist of images from three different domains, we decided to incorporate domain-shift into our approach. The goal is to train an unpaired domain-shift network, in order to create training examples from the domain of the otherwise unlabelled test domain, to generalize the classification and segmentation networks by learning robust features across domains. 

\begin{figure}[]
    \centering
    \includegraphics[width=\textwidth]{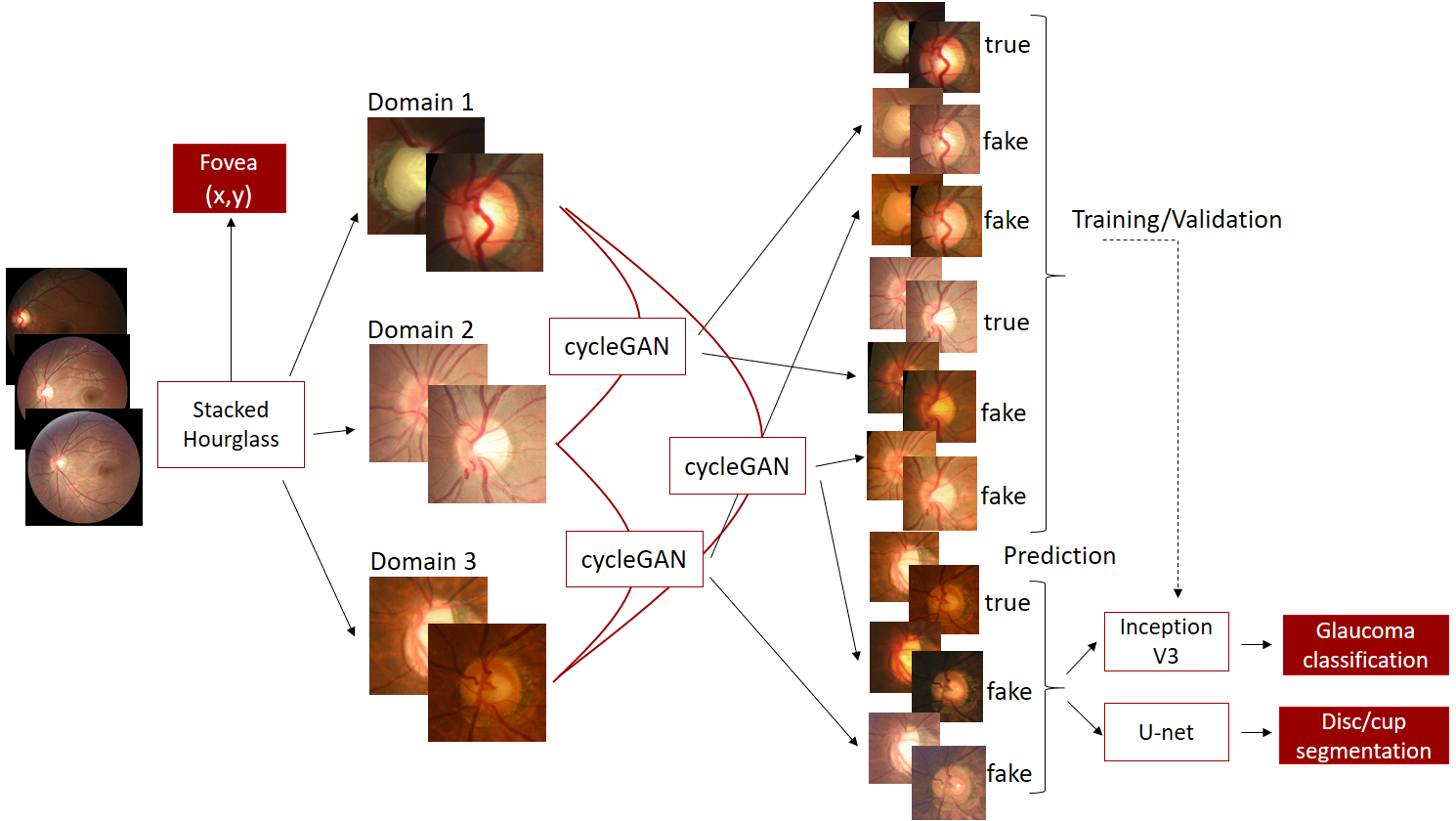}    
    \caption{Schematic representation of proposed method}
  \label{fig:pipeline}
\end{figure}

The full pipeline is shown in Figure \ref{fig:pipeline}. For the first step of ROI detection and fovea localization, we employ the stacked hourglass neural network trained on the original annotated training data from domain 1 and 2. The cropped input images in all three domains are then used to train three cycleGAN's for domain transfer across all three domains. An example of each image is artificially created in each domain, resulting in an increase in training and test data by a factor of 3. A combination of artificial domain-transferred images and original images are used to train an Inception V3 network for glaucoma classification and a U-net for optic disc and cup segmentation. All the steps of the proposed method will be described in details in the following sections.

The proposed method was implemented in Python using the deep learning framework Pytorch, and trained using a GeForce GTX 1070 8GB RAM graphics card.

\subsection{ROI detection and fovea localization}\label{sec:fovea}
Detecting the region-of-interest around the optic nerve head and locating fovea is carried out together using a stacked hourglass network \cite{Newell2016a} with two stacks. The network was originally proposed for human pose estimation and is chosen due to its ability to incorporate the interrelationship between the position of the optic nerve head and fovea. This setup allows the network to not only use image features for predicting the position of fovea and the optic nerve head, but also their relative position.
In the training images the center of the optic cup is found from the segmentation maps and a heatmap is created as a 2D gaussian placed at the optic cup center with a variance of 100 pixels. 
Similarly, a heatmap is created with a gaussian at the fovea location. 
The input image and heatmaps are resized to 256x256 and normalized to the range $[0,1]$. 
During training the images are augmented using the following transforms: random affine transformations with up to 20 degrees rotation, 100 pixels translation both vertical and horizontal and scaling with up to 20\%, color jitter which randomly changes hue ($\pm 10$), saturation ($[-0.2,0.5]$), and value ($\pm 0.3$), and horizontal and vertical flip. The transformation were applied with a probability of 0.5.

The network was trained using the Adam optimizer \cite{Kingma2015} with decreasing learning rate with a factor of 0.1 every 50th epoch. 
The initial learning rate was set to $0.001$, and trained with early stopping with a patience of 10 epochs.
The network makes use of spatial dropout with a dropout rate of 0.2. 
The batch size was set to 8 and all input images were shuffled during training. 

Location of fovea and center of optic cup are determined as pixel with the maximum intensity in each of the predicted heatmaps. 
The region of interest is constructed as 500x500 pixels cropped around the optic cup center. 

\subsection{Domain shift using cycleGAN}
To create the domain shifts between the image domains, several cycleGAN's were trained. The cycleGAN is a image-to-image translation model, where a mapping between an input and an output image is learned without the use of paired examples \cite{Zhu2017}. The cycleGAN models are trained using a group of images from the source domain and from the target domain, and the model learns to transfer images between these two domains. As seen in Figure \ref{fig:pipeline}, three cycleGAN's were needed to transfer between all combinations of the three domains. 
The standard implementation of cycleGAN was employed, with no modifications from the original implementation from \cite{Zhu2017}. 
No data augmentation was used, only the cropped region-of-interest from each of the three domains. 
The advantage of using cycleGAN is the fact that no labels are needed, as the domain shift is unpaired. 
The neural network will learn the representation of each domain without other information than the images. This is a big advantage for this application, where we have an unlabelled test dataset from a different domain than the training dataset. 
The results of domain transfer of three image examples can be seen in Figure \ref{tab:domainshift}. 
The images in the diagonal shows the original images from each of the domains, while the off-diagonal images show artificial images created by the cycleGAN. 
The figure shows that the cycleGAN creates artificial images that preserves the features of the original image (ie. the vessels and cup/disc size and shape) while incorporating features from the new domain (ie. colors). 

The neural networks for classification and segmentation were trained on the fixed training data, consisting of images from both domain 1 and 2. These images were transferred into each of the other domains, e.g. an image in domain 1 is transferred into both domain 2 and 3. The neural networks were therefore trained on image examples from all three domains, although many of them are artificially generated by the cycleGAN. This increases the size of the training dataset by a factor of 3, and ensures that the neural network learns the characteristics of the unlabelled test dataset.

\begin{table*}[ht!]
\begin{center}
\caption{Image examples from each of the three domains, and the domain-shifted images } \label{tab:domainshift}
\begin{tabular}{|c|c|c|c|}
\hline
 &  Domain 1 & Domain 2 & Domain 3 \\
\hline
\makecell{Image from \\ domain 1}         
&  \begin{minipage}{.2\textwidth}
      \includegraphics[width=\linewidth]{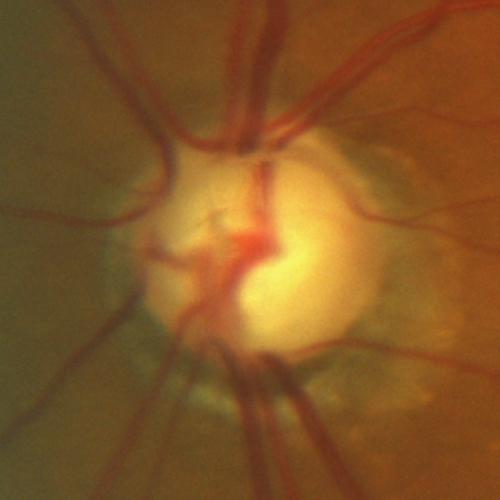}
    \end{minipage}
    & \begin{minipage}{.2\textwidth}
      \includegraphics[width=\linewidth]{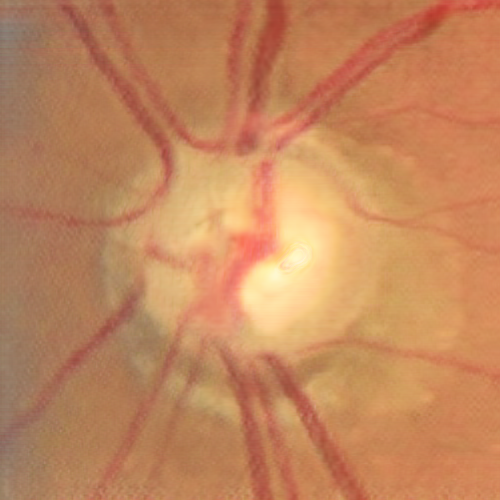}
    \end{minipage}
    &  \begin{minipage}{.2\textwidth}
          \includegraphics[width=\linewidth]{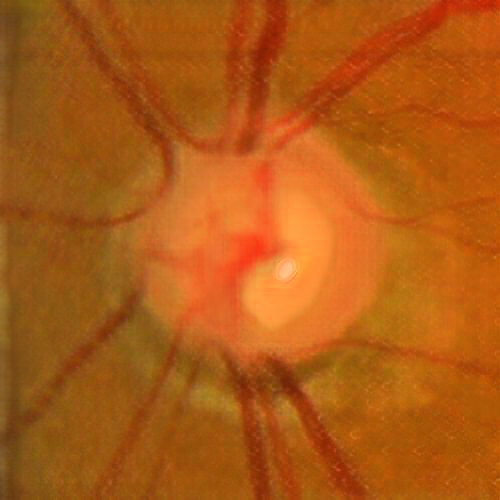}
        \end{minipage}
\\ \hline
    
\makecell{Image from \\ domain 2}           
&   \begin{minipage}{.2\textwidth}
      \includegraphics[width=\linewidth]{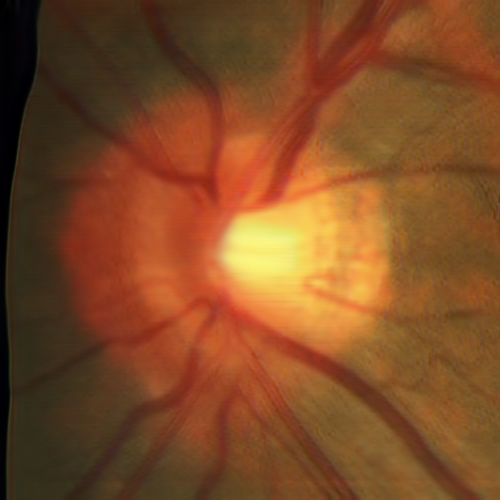}
    \end{minipage}
& \begin{minipage}{.2\textwidth}
      \includegraphics[width=\linewidth]{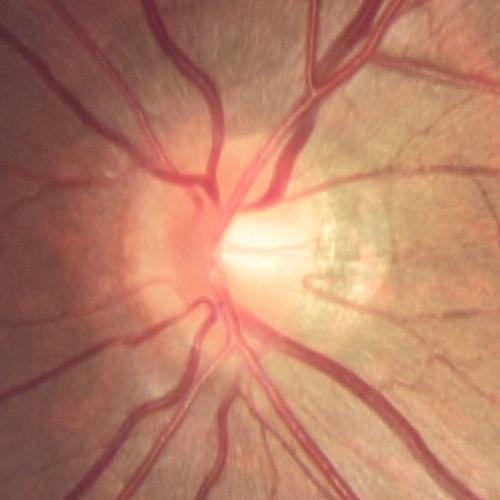}
    \end{minipage} 
&  \begin{minipage}{.2\textwidth}
      \includegraphics[width=\linewidth]{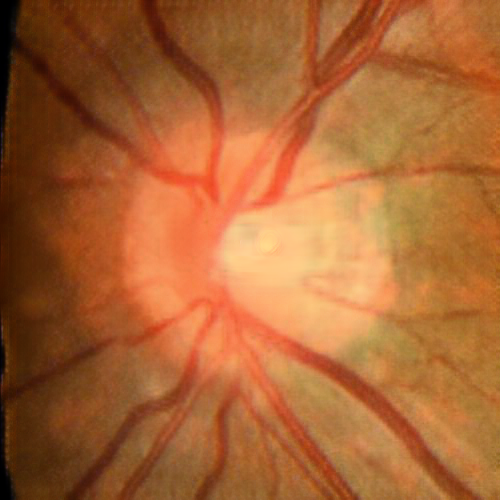}
    \end{minipage}
\\ \hline
\makecell{Image from \\ domain 3}          
&  \begin{minipage}{.2\textwidth}
      \includegraphics[width=\linewidth]{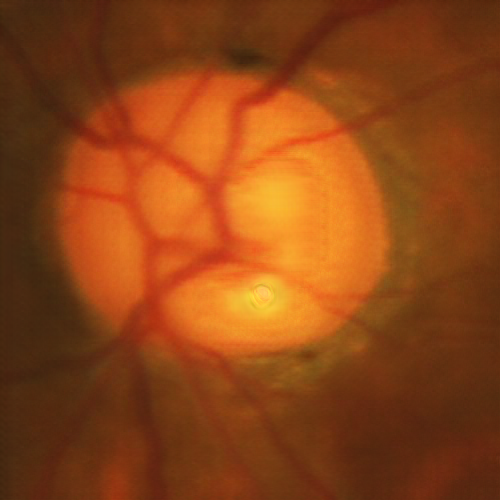}
    \end{minipage}
&  \begin{minipage}{.2\textwidth}
      \includegraphics[width=\linewidth]{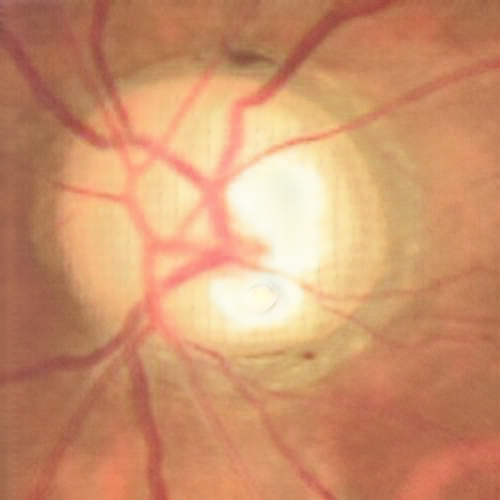}
    \end{minipage}
&  \begin{minipage}{.2\textwidth}
      \includegraphics[width=\linewidth]{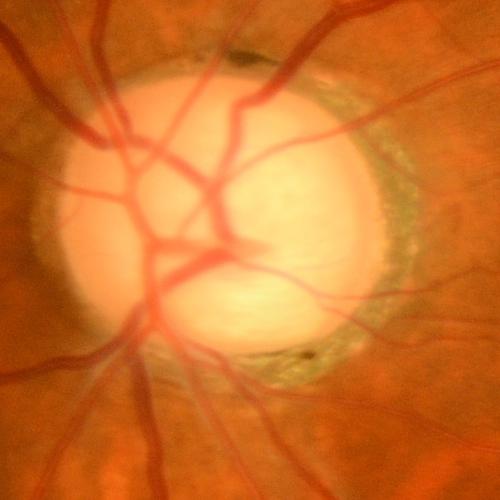}
    \end{minipage}
\\ \hline
\end{tabular}
\end{center}
\end{table*}

\subsection{Glaucoma classification}
Classification of glaucoma was performed using the Inception V3 network \cite{Szegedy2016a} for binary classification initialized with weights pre-trained on the ImageNet dataset. The weighted cross-entropy loss function was employed to enforce the network to learn to classify the under-represented class, glaucoma. The last layer of the neural network was changed to have the output dimensions two, to match the number of classes. Furthermore, the first half of the network (first four inception modules and the first grid size reduction) was frozen during training, and only the last half of the network was fine-tuned for the glaucoma classification task.

The input size for this network architecture is 299x299x3, as the images are RGB images. Thus the 500x500 input regions were resized to fit the network input. The network was trained using the Adam optimizer \cite{Kingma2015} with decreasing learning rate with a factor of 0.1 every eighth epoch. The initial learning rate was set to $10^{-4}$, and trained with early stopping with a patience of 10 epochs. The batch size was set to 60 and all input images were shuffled during training. 

Data augmentation was employed with a variation of transformations: random affine transformations with up to 20 degrees rotation, 60 pixels translation both vertical and horizontal, and scaling with up to 20\%, color jitter which randomly changes brightness ($\pm 10$), contrast ($\pm 10$), saturation ($\pm 10$), and hue ($\pm 10$), grey scale transformations, random perspective, and horizontal and vertical flips. The transformations were applied in a random order with a probability of 0.5. Before training, the images were also normalized with the standard parameters for the pre-trained Inception V3 network for Pytorch.

For prediction on the 400 test images, test-time augmentation was employed. In test-time augmentation, the test image is evaluated by the neural network several times with different transformations applied. 
The same transformations were used as during training, though only one transformation at a time, and each input image was evaluated with 10 different transformations. Besides the test-time augmentation, each image was also evaluated in each of the three domains.
The final prediction of glaucoma risk is computed by averaging the output of all 30 prediction (10 transformations in each of the three domains), and applying a softmax activation to obtain the probability for the two output classes.

\subsection{Optic disc and cup segmentation}
Segmentation of the optic disc an cup is carried out using a standard U-net \cite{Ronneberger2015b}. 
The input image and heatmaps are resized to 256x256 and normalized to the range $[0,1]$. 
During training the images are augmented using the following transforms: random affine transformations with up to 20 degrees rotation, 60 pixels translation both vertical and horizontal, and scaling with up to 20\%, color jitter which randomly changes hue ($\pm 10$), saturation ($[-0.2,0.5]$), and value ($\pm 0.3$) and horizontal and vertical flip. The transformations were applied with a probability of 0.5.

The network was trained using the Adam optimizer \cite{Kingma2015} with decreasing learning rate with a factor of 0.1 every 50th epoch. The initial learning rate was set to $0.001$, and trained with early stopping with a patience of 10 epochs. The batch size was set to 8 and all input images were shuffled during training. 

At prediction time, the image is evaluated 10 times in each of the three domains with different transformations similar to the augmentations used during training. 
In total the 30 different segmentation proposals are combined by averaging the network outputs before applying a softmax activation to obtain the final segmentation. 

\section{Challenge evaluation}
Our contribution was evaluated in the semi-final leaderboard with a overall ranking of 13 out of 22 contributions. For the classification task, our contribution ranked 13th with an AUC of 0.95. Optic disc and cup segmentation resulted in a mean cup dice of 0.86, a mean disc dice of 0.96 and cup-to-disc ratio relative mean error (CDR RME) of 0.04. On the segmentation task, our contribution ranked 11th out of 23 contributions. For fovea localization our contribution ranked 18th with a average euclidian distance of 29.7.
%
%
\bibliographystyle{splncs04}
%

\bibliography{EyeLoveGAN}

\end{document}